\documentclass[lettersize,journal]{IEEEtran}
\usepackage{amsmath,amsfonts}
\usepackage{algorithmic}
\usepackage{algorithm}
\usepackage{array}
\usepackage{xcolor}
\usepackage{float}
\usepackage{multirow}
\usepackage{booktabs}
\usepackage{placeins}
\usepackage{svg}
\usepackage{textcomp}
\usepackage{stfloats}
\usepackage{url}
\usepackage{verbatim}
\usepackage{graphicx}
\usepackage{cite}
\usepackage{academicons}
\usepackage{orcidlink}
\definecolor{orcidlogocol}{HTML}{A6CE39}
\usepackage{hyperref}

\title{Multi-Horizon Explainable Queue Forecasting and Resource-Aware Decision Support for Intelligent Border Control Using LightGBM}

\author{\IEEEauthorblockN{Varvara Mama\href{https://orcid.org/0009-0000-6369-7691}{\includegraphics[scale=0.35]{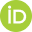}}, Fani D. Christidi  \href{https://orcid.org/0009-0008-7273-2856}{\includegraphics[scale=0.35]{ORCID-iD.png}}, Nikolaos C. Kapsalis\href{https://orcid.org/0000-0003-1084-2546}{\includegraphics[scale=0.35]{ORCID-iD.png}},Anargyros T. Baklezos\href{https://orcid.org/0000-0002-0532-6216}{\includegraphics[scale=0.35]{ORCID-iD.png}}, Member, \textit{IEEE}, \\and Christos D. Nikolopoulos\href{https://orcid.org/0000-0003-1344-4666}{\includegraphics[scale=0.35]{ORCID-iD.png}}, Member, \textit{IEEE}}.

\thanks{All authors are with the Department of Electronic Engineering,  Hellenic Mediterranean University; N. Kapsalis, C. D. Nikolopoulos, and A. T. Baklezos are also affiliated with the School of Electrical and Computer Engineering, National Technical University of Athens, Athens, Greece.}%
\thanks{This work extends the authors' conference study on multimodal AI for queue prediction and optimisation in intelligent border control systems.}
\thanks{Manuscript received April 19, 2021; revised August 16, 2021.}}

\begin{document}
\maketitle

\begin{abstract}
Border crossing points are stochastic service systems in which traffic demand, traveler categories, registration status, inspection requirements, staffing, and lane availability interact on multiple time scales. This paper extends an earlier multimodal queue-management framework by introducing an implementation-oriented, multi-horizon forecasting and decision-support architecture based on Light Gradient Boosting Machine (LightGBM) models. In contrast to a single forecasting block, the proposed approach constructs horizon-specific regressors for 15, 30, 60, and 120 min forecasts and jointly estimates future waiting time and queue lengths. The models operate on temporally lagged traffic, service, density, and queue features, while an operational layer incorporates lane compatibility, officer availability, possibly degraded resources, traveler category, preregistration information, and human-approval constraints. Quantile LightGBM models provide 95\% prediction intervals for waiting-time forecasts. A strict chronological hold-out protocol is adopted where models are trained exclusively on 28 days of synthetic five-minute-interval data and evaluated on a subsequent unseen seven-day period, preventing temporal leakage between model fitting and assessment. Across all forecast horizons, the proposed baseline obtains an overall waiting-time mean absolute error (MAE) of 6.438 min, root mean squared error (RMSE) of 13.079 min, and $R^2=0.9785$. Queue-length prediction achieves an overall MAE of 1.490 vehicles, RMSE of 2.558 vehicles, and $R^2=0.9977$. As expected, uncertainty and error increase with horizon; specifically, the waiting-time MAE rises from 5.007 min at 15 min to 8.472 min at 120 min. Beyond prediction, the framework produces machine-readable recommendations, alerts, aggregate key performance indicators, confidence information, and integration outputs suitable for border resource-management and decision-support systems. The results establish a reproducible synthetic-data baseline and a practical interface for subsequent validation with operational border-crossing data.
\end{abstract}


\section{Introduction}
\IEEEPARstart{B}{order} control is simultaneously a security process, a transportation process, and a constrained service system. The operational state of a border crossing point is determined not only by the number of arriving vehicles but also by vehicle class, traveler category, preregistration status, inspection requirements, staffing, equipment availability, and the instantaneous allocation of lanes. These factors create non-stationary and frequently bursty demand. They also create an inherent trade-off between minimizing delay and maintaining mandatory security and legal procedures.

The authors' previous work proposed a multimodal artificial-intelligence framework that combined heterogeneous border data, Long Short-Term Memory (LSTM) queue forecasting, Model Predictive Control (MPC), scheduling, and a Predictive Queue Management System (PQMS) \cite{ref1}. That work formulated border traffic as a dynamic multi-server queue and showed, in synthetic experiments, the potential of coupling prediction and optimization. The earlier architecture also identified real-time vehicle observations, historical data, preregistration information, weather, ferry information, and external traffic indicators as relevant data modalities.

The present work advances that concept from an architectural demonstration toward an implementation-oriented predictive service. Three limitations are addressed. First, operational decisions require several forecast horizons rather than a single next-step estimate. A 15-min prediction may support immediate lane decisions, whereas 60--120 min forecasts can support staffing and resource preparation. Second, a border decision-support service requires an explicit data contract that describes lanes, queue segments, operational constraints, uncertainty, recommendations, alerts, and downstream integration. Third, evaluation must prevent temporal leakage. Random train/test partitions are inappropriate for operational forecasting because future observations can indirectly inform training. We therefore use a strict chronological hold-out design.

The predictive core is changed from the LSTM emphasis of the precursor study to LightGBM \cite{ref2}. Gradient-boosted trees are attractive for an operational baseline because they naturally model nonlinear interactions among heterogeneous tabular features, require comparatively modest inference resources, and do not require the large training structured datasets often associated with deep sequence models. Recent traffic-forecasting research has also demonstrated the applicability of LightGBM to nonlinear spatio-temporal traffic data \cite{ref3}, while gradient boosting has a longer history in multi-step travel-time forecasting \cite{ref4}.

The principal contributions of this paper can be summarized as follows:
\begin{enumerate}
    \item a leakage-controlled multi-horizon forecasting formulation for border queues at 15, 30, 60, and 120 min (and can be scaled up to several hours);
    \item horizon-specific LightGBM regressors for waiting-time and queue-length increments, anchored to the current observed operational state;
    \item quantile-regression models for 95\% waiting-time prediction intervals;
    \item a queue-segment representation that distinguishes vehicle class, EU/Non-EU traveler category, and preregistered/unregistered flows;
    \item a resource-aware recommendation layer that checks lane compatibility, officer availability, equipment requirements, degraded lanes, and human-approval constraints;
    \item a machine-readable request/response interface that exposes predictions, alerts, recommendations, aggregate KPIs, quality information, and integration outputs; and
    \item a reproducible synthetic-data evaluation with a strict 28-day training and seven-day unseen test period.
\end{enumerate}

Unlike the precursor previous paper of the authors \cite{ref1}, the numerical results reported herein are not presented as a replacement for field validation. They constitute a controlled synthetic baseline for verifying the forecasting and software-integration methodology. This distinction is essential because synthetic queue dynamics are more regular than real border operations.

\section{Related Work and Research Context}

\subsection{Traffic and Queue Forecasting}
Traffic prediction has evolved from historical averages and classical time-series methods toward nonlinear machine-learning and deep-learning models. LSTM and bidirectional recurrent architectures have demonstrated the ability to capture temporal dependencies in traffic-flow sequences \cite{ref5,ref6}. Graph-based methods further exploit spatial interactions among road segments \cite{ref7,ref8}. These methods are powerful when large, structured spatio-temporal datasets are available.

Border crossings differ from conventional urban networks. The queue is generated by a controlled inspection process rather than by road capacity alone. Service rates can change abruptly when an officer is reassigned, equipment becomes unavailable, or an inspection lane enters a degraded state. Vehicle types also have substantially different service-time distributions. A bus carrying many passengers and a preregistered passenger car cannot be represented as equivalent service entities.

Tree-based gradient boosting offers a complementary approach. LightGBM uses histogram-based learning together with Gradient-based One-Side Sampling and Exclusive Feature Bundling to improve computational efficiency \cite{ref2}. Gradient boosting has previously been applied to multi-step travel-time prediction \cite{ref4}. More recently, LightGBM variants have been used for traffic-volume forecasting with temporal and spatial feature engineering \cite{ref3}. These properties motivate its use as a strong, computationally practical baseline for border decision-support systems.

\subsection{Prediction, Control, and Decision Support}
Prediction alone does not reduce queues. Forecasts must be translated into feasible interventions. MPC has been extensively investigated for traffic control because it optimizes future actions over a receding horizon while incorporating system constraints \cite{ref9,ref10}. Reinforcement-learning approaches have also been investigated for adaptive traffic control \cite{ref11,ref12}. In a border environment, however, fully autonomous actuation is not always appropriate. Lane reconfiguration and staffing decisions may require validation by authorized officers. Consequently, the proposed system is designed as human-in-the-loop decision support.

\subsection{Extension Beyond the Precursor Framework}
The precursor framework defined multimodal fusion, LSTM prediction, MPC/scheduling, and PQMS decision support. Its reported synthetic evaluation compared LSTM with historical-average and ARIMA baselines and reported operational improvements under simulated conditions. The present paper does not reuse those values as LightGBM results. Instead, it introduces a new dataset, a new predictor, a strict unseen-test protocol, horizon-dependent metrics, prediction intervals, and an explicit operational interface. Thus, the present contribution should be interpreted as an extension and implementation study rather than a re-reporting of the conference experiment.

\section{Border Queue System Model}

\subsection{Queue Segmentation}
Let $\mathcal{Q}=\{1,\ldots,Q\}$ denote queue segments. Each segment $q$ is characterized by
\begin{equation}
\chi_q=(c_q,r_q,v_q),
\end{equation}
where $c_q$ is traveler category, $r_q$ is registration status, and $v_q$ is vehicle type. In the implemented scenario the segments are EU preregistered cars, EU unregistered cars, Non-EU preregistered cars, Non-EU unregistered buses, and trucks/customs.

Let $Q_q(t)$ be the number of vehicles waiting in segment $q$ at time $t$. At a sampling interval $\Delta t=5$ min,
\begin{equation}
Q_q(t+\Delta t)=\max\{0,Q_q(t)+A_q(t)-S_q(t)\},
\label{eq:queue}
\end{equation}
where $A_q(t)$ is the number of arrivals and $S_q(t)$ is the effective number of completed services.

\subsection{Lane and Resource State}
Let $\mathcal{L}$ denote the set of lanes. Each lane $l$ has state
\begin{equation}
\ell_l(t)=
\{s_l,\mathcal{V}_l,\mathcal{C}_l,\mathcal{R}_l,n_l,\mathcal{E}_l\},
\end{equation}
where $s_l$ is lane status (e.g., open or degraded), $\mathcal{V}_l$ is the compatible set of vehicle types, $\mathcal{C}_l$ the allowed traveler categories, $\mathcal{R}_l$ the allowed registration states, $n_l$ the assigned officer count, and $\mathcal{E}_l$ the available equipment.

A binary compatibility variable is defined as
\begin{equation}
a_{ql}=
\begin{cases}
1, & v_q\in\mathcal{V}_l,\ c_q\in\mathcal{C}_l,\ r_q\in\mathcal{R}_l,\\
0, & \text{otherwise}.
\end{cases}
\end{equation}
Operational constraints additionally require specialized resources; for example, truck customs inspection may require a cargo scanner and bus processing may require group-processing capability.

\subsection{Forecasting Objective}
For forecast horizon $h\in\mathcal{H}$,
\begin{equation}
\mathcal{H}=\{15,30,60,120\}\ {\rm min},
\end{equation}
we estimate
\begin{equation}
\hat{W}_q(t+h),\qquad \hat{Q}_q(t+h),
\end{equation}
where $W_q$ is waiting time and $Q_q$ is queue length.

Rather than directly learning the absolute future state, the proposed model learns state increments:
\begin{align}
\Delta W_{q,h}(t) &= W_q(t+h)-W_q(t),\\
\Delta Q_{q,h}(t) &= Q_q(t+h)-Q_q(t).
\end{align}
The operational forecasts are reconstructed as
\begin{align}
\hat{W}_q(t+h)&=\max\{0,W_q(t)+\widehat{\Delta W}_{q,h}(t)\},\\
\hat{Q}_q(t+h)&=\max\{0,Q_q(t)+\widehat{\Delta Q}_{q,h}(t)\}.
\end{align}
This formulation anchors predictions to the latest measured state and reduces the risk that a synthetic absolute scale is transferred directly to a live operational state.

\section{Proposed Multi-Horizon LightGBM Framework}

\subsection{Input Features}
The baseline time-series record contains
\begin{equation}
\mathbf{x}_t =
[n_v,n_p,T_{\rm travel},\rho_v,\mu,Q,W,\mathbf{x}^{\rm lag},
\mathbf{x}^{\rm cal}],
\end{equation}
where $n_v$ is vehicle count per five-minute interval, $n_p$ is passenger count, $T_{\rm travel}$ is average travel time in seconds, $\rho_v$ is the traffic-density feature, $\mu$ is service rate, $Q$ is observed queue length in vehicles, $W$ is observed waiting time, $\mathbf{x}^{\rm lag}$ contains lagged queue/wait/service variables, and $\mathbf{x}^{\rm cal}$ contains cyclical/calendar descriptors.

For operational deployment, traffic density should be defined physically as
\begin{equation}
\rho_v(t)=\frac{N_v(t)}{L_{\rm obs}}\quad [{\rm vehicles/km}],
\end{equation}
where $N_v(t)$ is the number of vehicles observed on an approach segment of length $L_{\rm obs}$ km. In the current synthetic baseline, the density field is generated as a proxy congestion feature; conversion to a physically calibrated vehicles/km definition is therefore identified as a required step for pilot deployment.

\subsection{Horizon-Specific Gradient Boosting}
For each $h\in\mathcal{H}$, independent LightGBM regressors are trained:
\begin{align}
f_h^W &: \mathbf{x}_t \mapsto \widehat{\Delta W}_{h}(t),\\
f_h^Q &: \mathbf{x}_t \mapsto \widehat{\Delta Q}_{h}(t).
\end{align}
The gradient-boosting predictor can be represented as
\begin{equation}
F_M(\mathbf{x})=\sum_{m=1}^{M}\eta f_m(\mathbf{x}),
\end{equation}
where $f_m$ is the $m$th regression tree, $M$ is the number of boosting iterations, and $\eta$ is the learning rate. LightGBM's efficient histogram-based tree construction and sampling mechanisms make it suitable for repeated retraining and low-latency tabular inference \cite{ref2}.

Separate direct models are used instead of recursive one-step prediction. Recursive prediction can accumulate errors as the horizon grows. Direct horizon-specific models allow the mapping for 120 min to differ from the mapping for 15 min, which is useful when demand cycles and service-capacity changes operate on different time scales.

\subsection{Uncertainty Quantification}
Operational decisions require more than a point forecast. For each horizon, lower and upper waiting-time increments are estimated using quantile objectives:
\begin{align}
\widehat{\Delta W}_{h}^{(L)} &= f_{h,\alpha_L}^{W}(\mathbf{x}_t),\\
\widehat{\Delta W}_{h}^{(U)} &= f_{h,\alpha_U}^{W}(\mathbf{x}_t),
\end{align}
with $\alpha_L=0.025$ and $\alpha_U=0.975$. The resulting interval is
\begin{equation}
\mathcal{I}_{q,h}^{0.95}=
\left[
W_q(t)+\widehat{\Delta W}_{h}^{(L)},
W_q(t)+\widehat{\Delta W}_{h}^{(U)}
\right].
\end{equation}
The interval is reported to the decision-support layer alongside the point prediction. In operational validation, empirical coverage probability and interval width should be monitored to assess calibration.

\subsection{Operational Feature Adaptation}
The JSON request contains richer operational information than the baseline CSV. For each live queue segment, a feature adapter constructs a model input from current queue length, current waiting time, arrival rate, average service time, vehicle/passenger mix, assigned lanes, lane status, and external context. This adapter provides a controlled interface between learned time-series models and the operational data contract.

Importantly, fields without a direct training equivalent must not be silently treated as measured physical quantities. The software therefore records proxy mappings and uncertainty sources in the quality metadata. This makes the transition from synthetic to real data auditable.

\section{Resource-Aware Decision Layer}

\subsection{Risk Classification}
For queue $q$, a risk score is derived from predicted waiting time, predicted queue length, lane degradation, and forecast growth. A generic weighted score may be written as
\begin{equation}
R_q(h)=
\omega_W g_W(\hat W_q(t+h))+
\omega_Q g_Q(\hat Q_q(t+h))+
\omega_D d_q(t),
\end{equation}
where $g_W$ and $g_Q$ are normalized risk mappings and $d_q$ indicates degraded capacity. Thresholds map the score to low, medium, or high saturation/spillover risk.

\subsection{Weighted Waiting-Time Objective}
Let $w_q$ denote the operational priority weight of queue $q$. The resource-management objective is
\begin{equation}
J(t)=\sum_{q\in\mathcal{Q}} w_q \hat{W}_q(t+h) \hat{Q}_q(t+h).
\label{eq:weighted}
\end{equation}
A candidate lane assignment $y_{ql}\in\{0,1\}$ is feasible only if
\begin{align}
y_{ql} &\le a_{ql},\\
\sum_q y_{ql} &\le 1,\\
\sum_l n_l &\le N_{\rm officer},\\
y_{ql} &\le e_{ql},
\end{align}
where $e_{ql}$ represents required equipment/resource feasibility.

The current implementation uses a constrained recommendation layer rather than claiming a fully solved mixed-integer optimizer. This distinction is deliberate. It permits the forecasting service and the data interface to be validated before deploying an automated optimization policy.

\subsection{Human-in-the-Loop Actions}
Recommended actions are encoded with an action type, priority, affected lanes, affected queues, expected impact, checked constraints, explanation, and a Boolean human-approval requirement. This supports a governance model in which the algorithm proposes but authorized personnel approve safety- or security-relevant operational changes.

\section{Machine-Readable Interface}

\subsection{Request Contract}
The request structure contains: schema and module identifiers; checkpoint information; request timestamp; prediction horizons; operational context; lane states; current queue segments; real-time observations; external/calendar/weather/road features; resource constraints; and optimization preferences. This design separates the predictive model from upstream data providers.

The lane representation explicitly exposes allowed vehicle types, traveler categories, registration states, assigned officers, equipment, current queue length, vehicles in service, and estimated remaining service time. The queue-segment representation exposes passenger and vehicle counts, current waiting time, and assigned lanes.

\subsection{Response Contract}
The response contains model metadata, per-queue forecasts, confidence intervals, saturation/spillover risk, bottleneck flags, lane-allocation recommendations, alerts, aggregate KPIs, integration outputs, quality/uncertainty metadata, and execution status. Such a contract is important for integration because downstream systems should not depend on implementation-specific Python objects.

The architecture supports three principal consumers: a border resource-management component, a decision-support component, and a user interface. The same forecast can therefore be used for resource recommendations, explainable alerts, and dashboard visualization.

\section{Synthetic Dataset and Experimental Protocol}

\subsection{Synthetic Dataset}
A five-minute synthetic dataset was generated for 35 consecutive days. Five operational streams were represented: EU preregistered cars, EU unregistered cars, Non-EU preregistered cars, Non-EU unregistered buses, and trucks/customs. The simulation includes daily morning/evening demand peaks, weekly variability, weekend vehicle-class effects, incident-driven arrival surges, temporary service-capacity degradation, and class-specific service/passenger characteristics.

\begin{figure*}[t]
\centering
\includegraphics[width=0.94\textwidth]{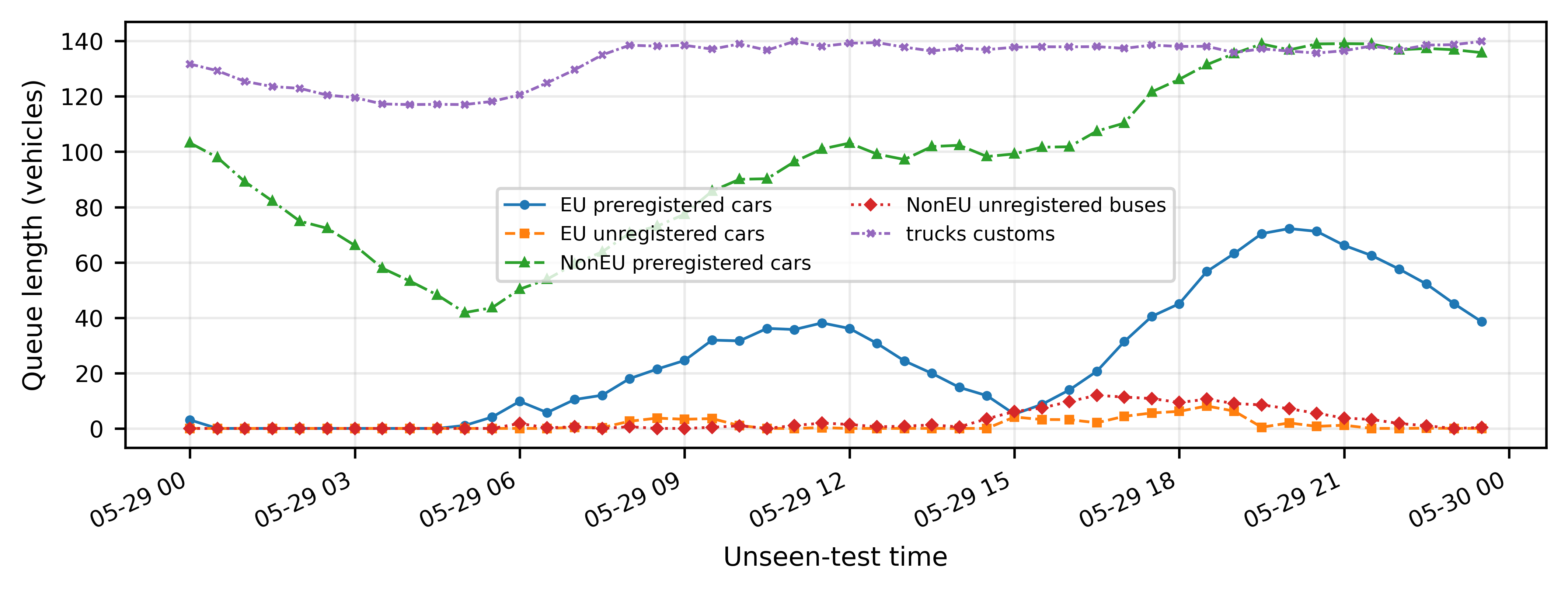}
\caption{Representative synthetic queue-length profiles for the five operational streams during one day of the unseen test period. The figure illustrates the heterogeneous temporal dynamics generated for cars, buses, and truck/customs processing. Thirty-minute plotting intervals are used for readability; the underlying data resolution is five minutes.}
\label{fig:profiles}
\end{figure*}

Figure~\ref{fig:profiles} illustrates the heterogeneous queue dynamics for a representative unseen-test day. The complete dataset contains 50,400 rows. The first 28 days form the training set (40,320 raw rows), and the final seven days form the unseen test set (10,080 raw rows). The split is chronological.

\begin{table}[t]
\caption{Synthetic Dataset Protocol}
\label{tab:data}
\centering
\begin{tabular}{ll}
\toprule
Sampling interval & 5 min\\
Total duration & 35 days\\
Queue streams & 5\\
Total observations & 50,400\\
Training period & First 28 days\\
Training observations & 40,320\\
Test period & Subsequent 7 days\\
Test observations & 10,080\\
Forecast horizons & 15, 30, 60, 120 min\\
\bottomrule
\end{tabular}
\end{table}

\subsection{Leakage-Controlled Hold-Out Evaluation}
For horizon $h$, the target is created by a group-wise temporal shift within each file. No target is allowed to cross the train/test boundary. The maximum training timestamp is verified to precede the minimum test timestamp. Models are fitted on training rows only. All reported MAE, RMSE, and $R^2$ values are calculated exclusively from the unseen test file.

This protocol is stricter than a random split because adjacent time-series observations are highly correlated. Random partitioning could place nearly identical temporal states in both sets and produce an optimistic estimate of deployment performance.

\subsection{Evaluation Metrics}
For $N$ test samples with observations $y_i$ and predictions $\hat y_i$,
\begin{equation}
{\rm MAE}=\frac{1}{N}\sum_{i=1}^{N}|y_i-\hat y_i|,
\end{equation}
\begin{equation}
{\rm RMSE}=
\sqrt{\frac{1}{N}\sum_{i=1}^{N}(y_i-\hat y_i)^2},
\end{equation}
and
\begin{equation}
R^2=1-\frac{\sum_i(y_i-\hat y_i)^2}
{\sum_i(y_i-\bar y)^2}.
\end{equation}
Waiting-time errors are reported in minutes and queue-length errors in vehicles.

\begin{algorithm}[t]
\caption{Strict Multi-Horizon Training and Evaluation}
\label{alg:training}
\begin{algorithmic}[1]
\REQUIRE Training file $\mathcal{D}_{tr}$, unseen test file $\mathcal{D}_{te}$, horizons $\mathcal{H}$
\STATE Verify $\max(t\in\mathcal{D}_{tr}) < \min(t\in\mathcal{D}_{te})$
\FOR{each $h\in\mathcal{H}$}
    \STATE Create $\Delta W_h,\Delta Q_h$ using shifts only within $\mathcal{D}_{tr}$
    \STATE Fit $f_h^W$, $f_h^Q$, and waiting-time quantile models on $\mathcal{D}_{tr}$
    \STATE Create unseen targets using shifts only within $\mathcal{D}_{te}$
    \STATE Predict $\hat W(t+h)$ and $\hat Q(t+h)$ on $\mathcal{D}_{te}$
    \STATE Compute MAE, RMSE, and $R^2$
\ENDFOR
\STATE Aggregate metrics across horizons
\end{algorithmic}
\end{algorithm}

\section{Results}

\subsection{Unseen Multi-Horizon Performance}

\begin{figure*}[t]
\centering
\includegraphics[width=0.94\textwidth]{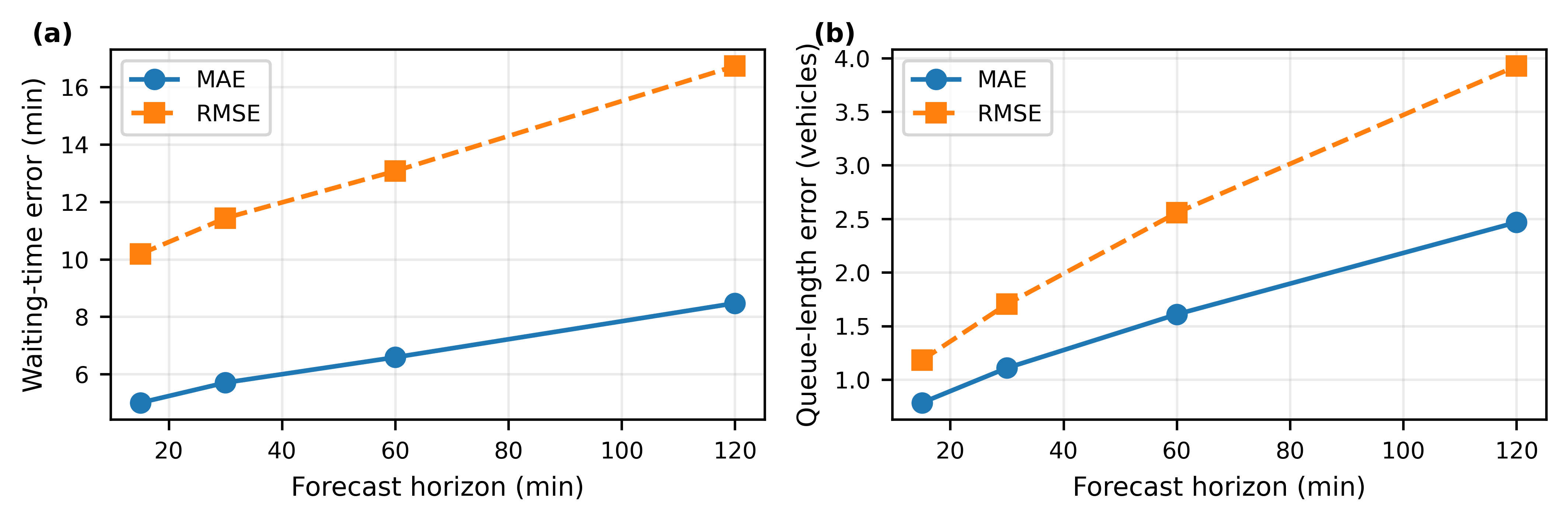}
\caption{Unseen-test prediction error as a function of forecast horizon: (a) waiting-time MAE and RMSE; (b) queue-length MAE and RMSE. Both error measures increase with horizon, illustrating the progressive uncertainty of longer-range border-queue forecasting.}
\label{fig:horizon_errors}
\end{figure*}

Table~\ref{tab:results} reports the strict hold-out results, while Fig.~\ref{fig:horizon_errors} visualizes the growth of MAE and RMSE with forecast horizon. At 15 min, waiting-time prediction achieves an MAE of 5.007 min and $R^2=0.9870$, while queue-length prediction achieves an MAE of 0.784 vehicles and $R^2=0.9995$. At 120 min, waiting-time MAE increases to 8.472 min and queue-length MAE to 2.468 vehicles. This monotonic deterioration is consistent with increasing forecast uncertainty.

\begin{table*}[t]
\caption{Unseen Test Performance by Forecast Horizon}
\label{tab:results}
\centering
\begin{tabular}{c c c c c c c}
\toprule
Horizon & \multicolumn{3}{c}{Waiting Time} & \multicolumn{3}{c}{Queue Length}\\
\cmidrule(lr){2-4}\cmidrule(lr){5-7}
(min) & MAE (min) & RMSE (min) & $R^2$ & MAE (veh.) & RMSE (veh.) & $R^2$\\
\midrule
15  & 5.007 & 10.186 & 0.9870 & 0.784 & 1.182 & 0.9995\\
30  & 5.705 & 11.435 & 0.9836 & 1.108 & 1.704 & 0.9990\\
60  & 6.588 & 13.074 & 0.9785 & 1.609 & 2.557 & 0.9977\\
120 & 8.472 & 16.729 & 0.9648 & 2.468 & 3.926 & 0.9946\\
\midrule
Overall & 6.438 & 13.079 & 0.9785 & 1.490 & 2.558 & 0.9977\\
\bottomrule
\end{tabular}
\end{table*}

The overall waiting-time RMSE is approximately twice the MAE. This indicates that a minority of difficult states contributes disproportionately to squared error, as expected during abrupt demand or capacity changes. Such states are operationally important because they correspond to precisely the periods in which congestion-management support is most valuable.

\begin{figure*}[t]
\centering
\includegraphics[width=0.94\textwidth]{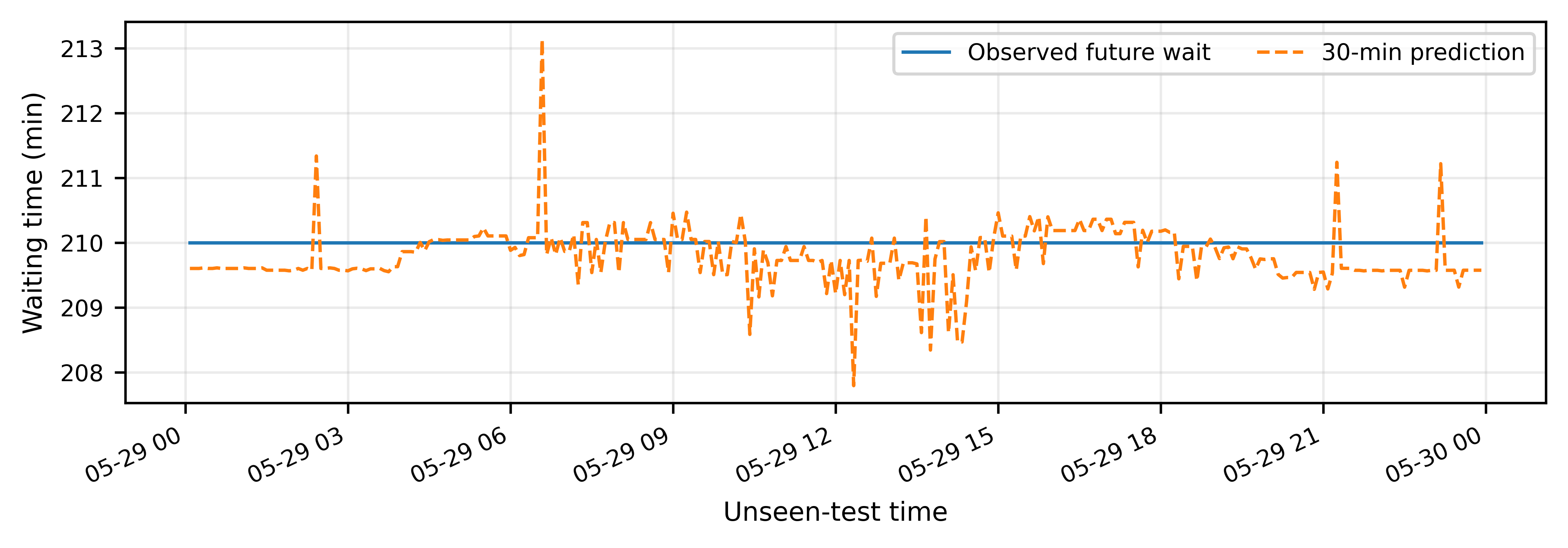}
\caption{Representative 30-min-ahead waiting-time forecast for the \texttt{trucks\_customs} stream during one day of the strictly unseen test period. The graph compares the future observed waiting time with the LightGBM prediction and illustrates both normal tracking and transient forecast deviations.}
\label{fig:timeseries}
\end{figure*}

\begin{figure}[t]
\centering
\includegraphics[width=\columnwidth]{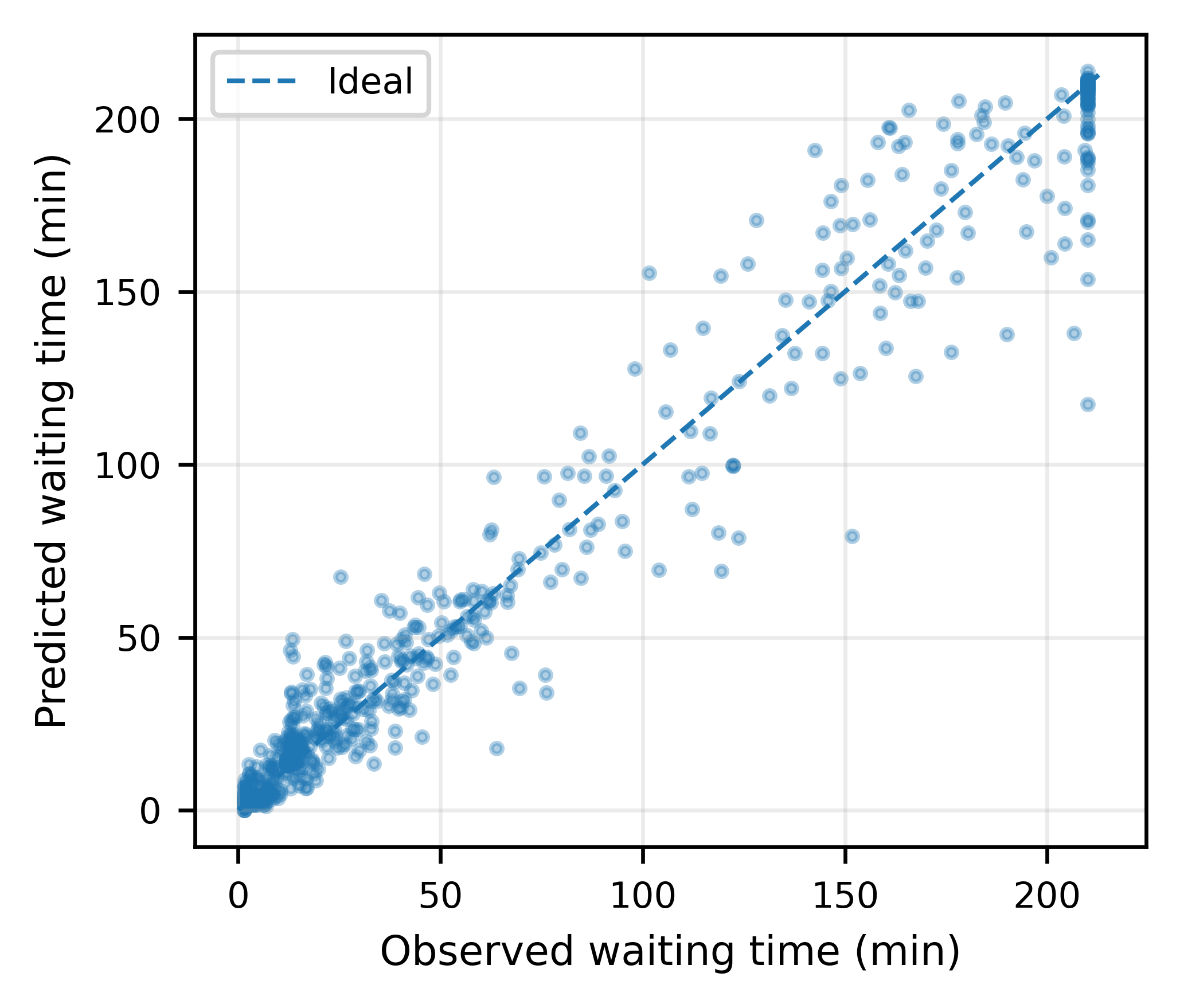}
\caption{Predicted versus observed 30-min-ahead waiting time across all queue streams in the unseen test set. Every tenth point is displayed for visual clarity; all test samples are retained in the numerical metrics.}
\label{fig:scatter}
\end{figure}

\subsection{Horizon Dependence}
Figure~\ref{fig:timeseries} provides a representative 30-min-ahead truck/customs trajectory, and Fig.~\ref{fig:scatter} shows the corresponding observed-versus-predicted relationship across all queue streams. Relative to the 15-min forecast, waiting-time MAE increases by approximately 69\% at 120 min. Queue-length MAE increases by more than a factor of three. Nevertheless, $R^2$ remains above 0.96 for waiting time and above 0.99 for queue length at all four horizons in the synthetic experiment.

These high coefficients of determination should not be interpreted as expected field performance. The synthetic generator creates internally coherent relationships between arrivals, service capacity, queue accumulation, and waiting time. Real border data will include unobserved operational decisions, measurement errors, heterogeneous inspection outcomes, behavioral effects, and potentially missing or delayed signals. The present results therefore validate the pipeline under controlled conditions rather than establish operational accuracy.

\subsection{Operational Output Example}
When the trained forecasting pipeline is applied to the example request, the response identifies several queues with elevated forecast risk and generates resource-aware actions. The generated operational output includes total vehicles waiting, mean and maximum 30-min predicted waiting time, weighted waiting-time KPI, bottleneck count, open/degraded lane counts, and an overall operational-risk label. These quantities are derived from the live request state and should not be confused with the unseen-test accuracy metrics in Table~\ref{tab:results}.

\subsection{Ablation Study}

\begin{figure*}[t]
\centering
\includegraphics[width=0.94\textwidth]{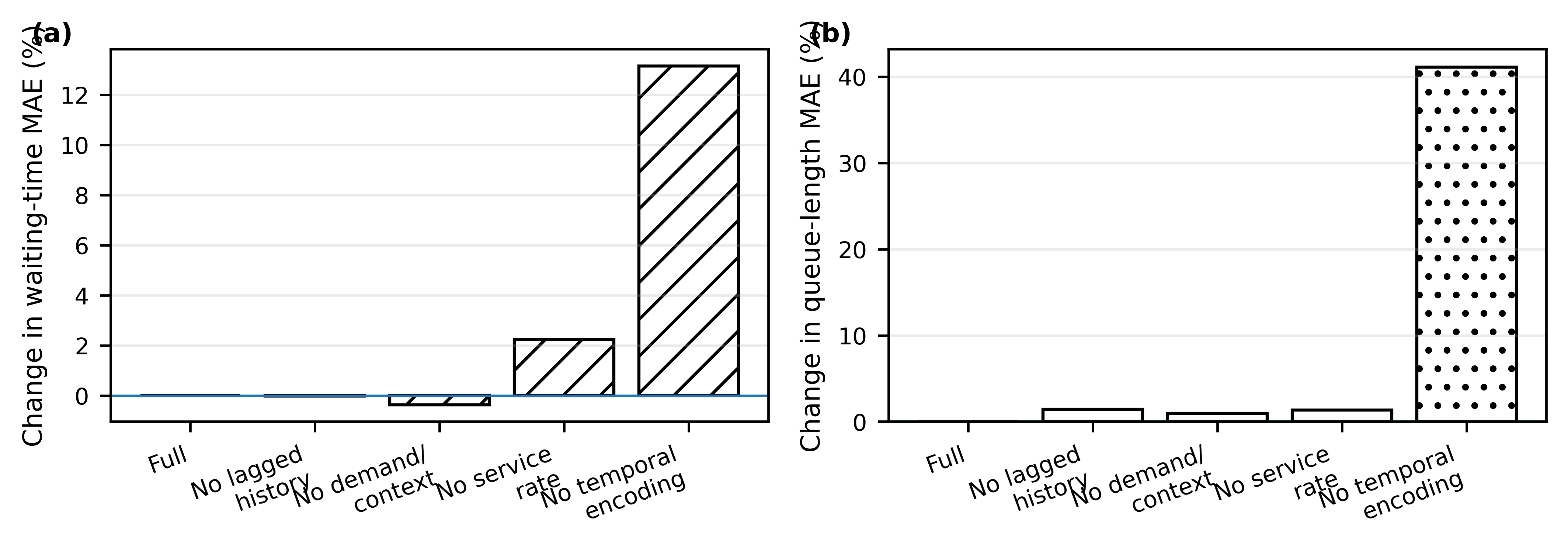}
\caption{Relative change in unseen-test MAE under feature ablation, referenced to the full LightGBM model: (a) waiting-time prediction and (b) queue-length prediction. Removing cyclical temporal encoding produces the largest degradation, particularly for queue-length forecasting.}
\label{fig:ablation}
\end{figure*}

To quantify the contribution of the principal feature groups, an ablation study was performed under exactly the same strict chronological hold-out protocol used for the full model. In each experiment, the corresponding feature group was removed from both training and unseen-test inference; all LightGBM hyperparameters, horizons, targets, and train/test periods were otherwise unchanged. Table~\ref{tab:ablation} reports metrics aggregated across the four horizons.

\begin{table*}[t]
\caption{Feature Ablation Study Under the Same Unseen Hold-Out Protocol}
\label{tab:ablation}
\centering
\begin{tabular}{l c c c c c c}
\toprule
Configuration & \multicolumn{3}{c}{Waiting Time} & \multicolumn{3}{c}{Queue Length}\\
\cmidrule(lr){2-4}\cmidrule(lr){5-7}
 & MAE (min) & RMSE (min) & $R^2$ & MAE (veh.) & RMSE (veh.) & $R^2$\\
\midrule
Full model & 6.438 & 13.079 & 0.9785 & 1.490 & 2.558 & 0.9977\\
Without lagged history & 6.436 & 13.078 & 0.9785 & 1.511 & 2.599 & 0.9976\\
Without demand/context & 6.414 & 13.048 & 0.9786 & 1.504 & 2.573 & 0.9977\\
Without service rate & 6.582 & 13.407 & 0.9774 & 1.510 & 2.576 & 0.9977\\
Without temporal encoding & 7.284 & 14.542 & 0.9734 & 2.102 & 3.937 & 0.9946\\
\bottomrule
\end{tabular}
\end{table*}

The relative degradation is visualized in Fig.~\ref{fig:ablation}. The strongest deterioration occurs when the cyclical time-of-day variables are removed. Waiting-time MAE increases from 6.438 to 7.284 min (approximately 13.1\%), while queue-length MAE increases from 1.490 to 2.102 vehicles (approximately 41.1\%). Queue RMSE increases by approximately 53.9\%. This result indicates that the synthetic border process contains strong recurring temporal structure that cannot be reconstructed completely from the instantaneous queue state.

Removing the service-rate feature produces the second clearest waiting-time degradation: MAE rises to 6.582 min and RMSE to 13.407 min. This is physically consistent with the queue balance in (\ref{eq:queue}), because effective processing capacity determines whether arriving demand is absorbed or accumulated.

In contrast, removing lagged history produces only a negligible change in waiting-time error and a small degradation in queue-length error. More notably, removing the demand/context group (vehicle count, passenger count, average travel time, density, and vehicle-arrival lag features) slightly improves aggregate waiting-time MAE from 6.438 to 6.414 min, while marginally degrading queue MAE. This result must not be interpreted as evidence that demand/context data are unnecessary in real border operations. Rather, it exposes a property of the present synthetic generator: the current observed queue and waiting state, service capacity, and cyclical time variables already encode much of the generated system dynamics, leaving some demand variables redundant. The ablation therefore provides an important warning against over-interpreting synthetic feature importance. Real pilot data are expected to contain delayed effects, measurement noise, exogenous disturbances, and operational interventions for which multimodal demand/context features may become substantially more informative.

The ablation findings are consistent at a methodological level with recent LightGBM traffic research showing that explicit temporal structure can materially affect forecasting performance. Wang \emph{et al.} incorporated cyclical temporal features and dynamic temporal weighting in a LightGBM framework and reported strong gains over standard machine-learning and recurrent baselines \cite{ref3}. The present study reaches a related conclusion through removal experiments: temporal representation is the most consequential tested feature group for the synthetic border queue process.

\subsection{Methodological Comparison with Related Studies}
Table~\ref{tab:literature} positions the proposed framework relative to selected peer-reviewed traffic and multi-horizon forecasting studies. The table is intended as a methodological comparison rather than a numerical leaderboard. The datasets, targets, sampling intervals, normalization procedures, and prediction horizons differ substantially; consequently, MAE or RMSE values from different rows are \emph{not directly comparable}.

\begin{table*}[t]
\caption{Positioning Against Selected Published Forecasting Studies}
\label{tab:literature}
\centering
\footnotesize
\begin{tabular}{p{2.7cm} p{2.0cm} p{2.3cm} p{2.2cm} p{2.4cm} p{3.1cm}}
\toprule
Study & Application & Models/approach & Forecast setting & Reported result & Relation to present work\\
\midrule
Zhang and Haghani \cite{ref4}
& Freeway travel time
& Gradient boosting regression trees; ensemble/statistical baselines
& Multi-step ahead
& GBM reported advantageous prediction accuracy and interpretability
& Establishes gradient boosting as a suitable nonlinear, interpretable travel-time predictor.\\

Berliotti \emph{et al.} \cite{ref15}
& Road traffic flow, 14 sensors
& LightGBM, XGBoost, CatBoost, Random Forest
& Hourly traffic; two unseen one-week periods
& First test week: LightGBM MAE 0.0679; XGBoost 0.0649; CatBoost 0.0637 (normalized task scale)
& Demonstrates that no boosting method is universally dominant and motivates same-data model comparison.\\

Wang \emph{et al.} \cite{ref3}
& 209 urban signalized intersections
& Dynamically Weighted LightGBM versus LightGBM, XGBoost, LSTM, GRU, RF, SVM, KNN
& Hourly traffic-volume forecasting
& DW-LGBM RMSE 5.54, $R^2=0.9996$; standard LightGBM RMSE 118.13, $R^2=0.6574$
& Supports temporal weighting/encoding and shows the importance of feature architecture around LightGBM.\\

Hari \emph{et al.} \cite{ref16}
& Multi-location pedestrian/crowd demand
& LightGBM, CatBoost, LSTM and other models
& Multiple prediction windows
& Best model changes with location and horizon; reported $R^2$ ranges from negative values at difficult low-volume sites to $>0.5$ at busier sites
& Reinforces horizon- and location-dependent evaluation rather than reliance on one aggregate metric.\\

Present work
& Border queue waiting time and queue length
& Horizon-specific LightGBM + quantile LightGBM + resource-aware decision layer
& 15/30/60/120 min; strict 28-day train/7-day unseen test
& Wait: MAE 6.438 min, RMSE 13.079 min, $R^2=0.9785$; Queue: MAE 1.490 veh., RMSE 2.558 veh., $R^2=0.9977$
& Adds border-specific segmentation, service/resource constraints, uncertainty intervals, JSON interoperability, and operational recommendations.\\
\bottomrule
\end{tabular}
\end{table*}

The literature comparison highlights three points. First, gradient-boosted trees are established predictors for transportation variables, including multi-step travel time \cite{ref17}. Second, published real-world traffic studies show that LightGBM does not automatically dominate other ensemble learners. 
Berlotti \emph{et al.} found XGBoost and CatBoost competitive with or better than LightGBM in their real-world traffic-sensor experiment \cite{ref18}. Therefore, the high $R^2$ values obtained in the present synthetic experiment cannot, by themselves, support a claim of algorithmic superiority.

Third, recent results indicate that the representation of temporal structure can matter as much as the underlying boosting algorithm. Wang \emph{et al.} reported a large difference between standard LightGBM and a dynamically weighted LightGBM architecture on real intersection data \cite{ref19}. This observation is particularly relevant to the present ablation results, where removing time-of-day encoding causes the largest deterioration. Published multi-horizon crowd forecasting similarly shows that the preferred model can change with forecast window and location \cite{ref16}. Accordingly, future border evaluation should report performance separately by horizon, checkpoint, queue category, and operational regime.

The present framework differs from these journal studies in its target and system boundary. It predicts both a service-system state (queue length) and a traveler-facing outcome (waiting time), and couples the forecasts to explicit lane, staffing, equipment, compatibility, and human-approval constraints. Its main novelty is therefore not a claim that standard LightGBM universally outperforms published traffic predictors, but the integration of multi-horizon boosting, uncertainty estimation, border-specific queue semantics, and resource-aware decision support in a leakage-controlled operational pipeline.

\subsection{Comparison With the Precursor Study}
The conference study emphasized LSTM-based queue prediction and reported synthetic comparisons with ARIMA and historical-average baselines. The current experiment uses a newly generated dataset and a different evaluation protocol; therefore, direct numerical comparison between the old LSTM table and the new LightGBM table would not be scientifically valid. A fair LSTM-versus-LightGBM comparison requires both algorithms to be retrained on exactly the same 28-day training set and evaluated on exactly the same seven-day hold-out set. This is identified as the next comparative experiment.

\section{Discussion}

\subsection{Why LightGBM Is Appropriate for the Operational Baseline}
The input to a border-control forecasting service is predominantly structured tabular data: counts, service rates, lane status, queue state, time variables, categorical descriptors, and external context. LightGBM can model nonlinear interactions among these variables without requiring a recurrent hidden state. This simplifies training, deployment, and model inspection.

The approach is also modular. A separate model exists for each horizon, so a poor 120-min model does not directly contaminate the 15-min estimate. Conversely, this increases the number of trained models. In a production environment, model registry, versioning, monitoring, and retraining must therefore be managed explicitly.

\subsection{Prediction Intervals and Decision Risk}
Point forecasts can create false precision. If two candidate lane actions have similar expected benefit but one is associated with a much wider waiting-time interval, the uncertainty should influence the recommendation. A risk-aware extension of (\ref{eq:weighted}) is
\begin{equation}
J_{\rm risk}=J+\gamma\sum_q w_q
\left(U_{q,h}-L_{q,h}\right),
\end{equation}
where $L_{q,h}$ and $U_{q,h}$ are interval bounds and $\gamma$ controls uncertainty aversion. This formulation provides a direct route for coupling probabilistic forecasting to resource optimization.

\subsection{Explainability}
Tree ensembles support several levels of explanation. Global feature importance can identify variables that dominate model behavior, while local attribution methods can explain a specific queue forecast. For operational use, explanations should be translated into domain language, e.g., ``truck waiting time is increasing because arrival intensity has exceeded effective service capacity while one customs lane is degraded.'' The response contract already contains an explanation field, allowing more advanced attribution methods to be integrated without changing downstream interfaces.

\subsection{Data Semantics}
A critical deployment issue is semantic consistency. Vehicle count is an interval flow measure, queue length is a stock measured in vehicles, service rate is a capacity measure, and traffic density should be a spatial concentration measured in vehicles/km. Mixing these quantities or using undocumented proxy features can undermine both reproducibility and operational trust. Accordingly, the pilot dataset should include a formal data dictionary with units, sampling windows, sensor origin, missing-value conventions, and quality flags.

\subsection{Relationship to MPC and Scheduling}
The present implementation focuses on forecasting and constrained recommendations. It remains compatible with the MPC concept introduced in the precursor work. The LightGBM forecast can supply exogenous demand and queue trajectories to an MPC or mixed-integer scheduling layer. A receding-horizon controller could solve
\begin{equation}
\min_{\mathbf{u}_{t:t+H}}
\sum_{k=1}^{H}
\left[
\mathbf{w}^{T}\hat{\mathbf{Q}}(t+k)
+\lambda\|\Delta\mathbf{u}(t+k)\|_1
\right],
\end{equation}
subject to lane, staffing, equipment, separation, and security constraints. The first feasible action is presented for human approval, and the optimization is repeated when new observations arrive.

\section{Limitations and Threats to Validity}
The main limitation is the synthetic nature of the dataset. The generator intentionally creates physically plausible queue dynamics, but its equations also induce strong learnable relationships. The resulting $R^2$ values are consequently expected to exceed those obtainable in heterogeneous operational environments.

Second, the present experiment does not yet perform a same-data comparison against LSTM, ARIMA, XGBoost, or naive persistence. Such baselines are required before claiming that LightGBM is superior. The current contribution is a rigorous LightGBM baseline and interface implementation.

Third, the 95\% intervals are generated using quantile regression, but empirical coverage calibration has not yet been reported. Operational validation should report prediction interval coverage probability, mean interval width, and coverage under peak/disruption regimes.

Fourth, the synthetic density variable is presently a congestion proxy. Before pilot deployment, it should be replaced by a calibrated vehicles/km measurement tied to a defined road-segment length.

Finally, the recommendation engine is constraint-aware but is not yet a complete mathematical optimizer. Expected impact estimates must ultimately be validated through discrete-event simulation or field data before being used as quantitative claims.

\section{Future Experimental Methodology}
The next validation stage should use the same chronological split to compare persistence, historical average, ARIMA, LSTM, XGBoost, and LightGBM. Statistical significance can be assessed using paired forecast-error tests across common timestamps. Evaluation should be stratified by queue type, demand regime, incident status, and forecast horizon.

A second stage should evaluate uncertainty calibration. A third should couple forecasts to a discrete-event border simulator and measure operational KPIs such as average waiting time, maximum waiting time, throughput, officer utilization, lane utilization, and spillover probability under identical demand scenarios. Finally, real pilot data should replace the synthetic generator while retaining the same JSON contract and evaluation code. This progression separates software verification, predictive validation, control-policy validation, and operational field validation.

\section{Conclusion}
This paper presented an implementation-oriented extension of a multimodal intelligent border queue-management framework. The proposed approach replaces the single-model emphasis of the precursor study with direct multi-horizon LightGBM forecasting at 15, 30, 60, and 120 min, strict chronological hold-out evaluation, quantile-based uncertainty estimation, explicit queue segmentation, resource-aware recommendations, and a machine-readable integration interface.

On a controlled 35-day synthetic dataset, with 28 days used exclusively for training and seven subsequent days reserved for unseen testing, the method achieved an overall waiting-time MAE of 6.438 min, RMSE of 13.079 min, and $R^2=0.9785$. Queue-length forecasting achieved an MAE of 1.490 vehicles, RMSE of 2.558 vehicles, and $R^2=0.9977$. Forecast error increased with prediction horizon, emphasizing the importance of horizon-specific performance reporting and uncertainty information.

The principal value of the present results is methodological: the work establishes a reproducible, leakage-controlled baseline and a practical bridge between predictive analytics and border resource-management/decision-support systems. Validation on operational data, same-dataset comparisons with LSTM and statistical baselines, calibrated uncertainty analysis, and closed-loop optimization remain necessary before deployment claims can be made.

\section*{Acknowledgment}
This research was partially funded by the European Commission under the Horizon Europe Programme through the ORION project under Grant Agreement No. 101225611. Views and opinions expressed are those of the authors only and do not necessarily reflect those of the European Union.

\end{document}